%% file: emnlp2021.tex
\pdfoutput=1

\documentclass[11pt]{article}

\usepackage{emnlp2021}

\usepackage{times,amsmath}
\usepackage{latexsym,multicol,multirow,booktabs}
\usepackage{euflag}

\usepackage[T1]{fontenc}

\usepackage[utf8]{inputenc}

\usepackage{microtype}

\usepackage{todonotes}
\makeatletter
\newcommand*\iftodonotes{\if@todonotes@disabled\expandafter\@secondoftwo\else\expandafter\@firstoftwo\fi}
\makeatother

\usepackage{cleveref}
\crefformat{section}{\S#2#1#3}
\crefformat{subsection}{\S#2#1#3}
\crefformat{subsubsection}{\S#2#1#3}
\crefrangeformat{section}{\S\S#3#1#4 to~#5#2#6}
\crefmultiformat{section}{\S\S#2#1#3}{ and~#2#1#3}{, #2#1#3}{ and~#2#1#3}
\usepackage{refstyle}
\crefformat{figure}{#2Fig.~#1#3}
\crefmultiformat{figure}{Figs.~#2#1#3}{ and~#2#1#3}{, #2#1#3}{ and~#2#1#3}
\crefformat{table}{#2Tab.~#1#3}
\crefmultiformat{table}{Tabs.~#2#1#3}{ and~#2#1#3}{, #2#1#3}{ and~#2#1#3}
\crefformat{equation}{#2Eq.~(#1#3)}
\crefformat{appendix}{#2App.~\S#1#3}

\usepackage{xcolor}
\usepackage{graphicx}
\usepackage{amsfonts, amsmath, amsthm, amssymb}
\usepackage{xspace}
\usepackage{float}
\usepackage{multirow}
\usepackage{enumitem}
\usepackage{booktabs}
\usepackage{comment}
\usepackage[export]{adjustbox}
\usepackage{array}
\usepackage{xcolor}
\usepackage{soul}
\usepackage{caption}
\usepackage{subcaption}
\usepackage{tabularx}
\usepackage[normalem]{ulem}
\useunder{\uline}{\ul}{}

\title{On Language Models for Creoles}

\usepackage{tipa}
\newcommand{\wuhan}{\text{\normalfont \textipa{W}}}
\newcommand{\ku}{\text{\normalfont  \textipa{C}}}
\newcommand{\uu}{\text{\normalfont \textipa{U}}}
\newcommand{\kul}{\text{\normalfont \textipa{D}}}

\author{Heather Lent$^{\ku}$ \ \ Emanuele Bugliarello$^{\ku}$ \ \ Miryam de Lhoneux$^{\ku,\uu,\kul}$\\ {\bf Chen Qiu$^{\wuhan}$ \ \ Anders Søgaard$^\ku$} \\
         $^\ku$ University of Copenhagen, Denmark  \\
          $^\uu$ Uppsala University, Sweden
          $^\kul$ KU Leuven, Belgium\\
          $^\wuhan$ Wuhan University of Science and Technology, China \\
           \texttt{\{hcl, emanuele, ml, soegaard\}@di.ku.dk}\\ \texttt{chen@wust.edu.cn} \\
           }

\begin{document}
\maketitle
\begin{abstract}
Creole languages such as Nigerian Pidgin English and Haitian Creole are under-resourced and largely ignored in the NLP literature. Creoles typically result from the fusion of a foreign language with multiple local languages, and what grammatical and lexical features are transferred to the creole is a complex process \cite{Sessarego2020NotAG}. While creoles are generally stable, the prominence of some features may be much stronger with certain demographics or in some linguistic situations \cite{winford99variation,Patrick1999}. This paper makes several contributions: We collect existing corpora and release models for Haitian Creole, Nigerian Pidgin English, and Singaporean Colloquial English. We evaluate these models on intrinsic and extrinsic tasks. Motivated by the above literature, we compare standard language models with distributionally robust ones and find that, somewhat surprisingly, the standard language models are superior to the distributionally robust ones. We investigate whether this is an effect of over-parameterization or relative distributional stability, and find that the difference persists in the absence of over-parameterization, and that drift is limited, confirming the relative stability of creole languages.  

\end{abstract}

\section{Introduction}
\begin{figure}
    \centering
    \includegraphics[width=\linewidth, trim={0cm 0cm 0cm 0cm}, clip]{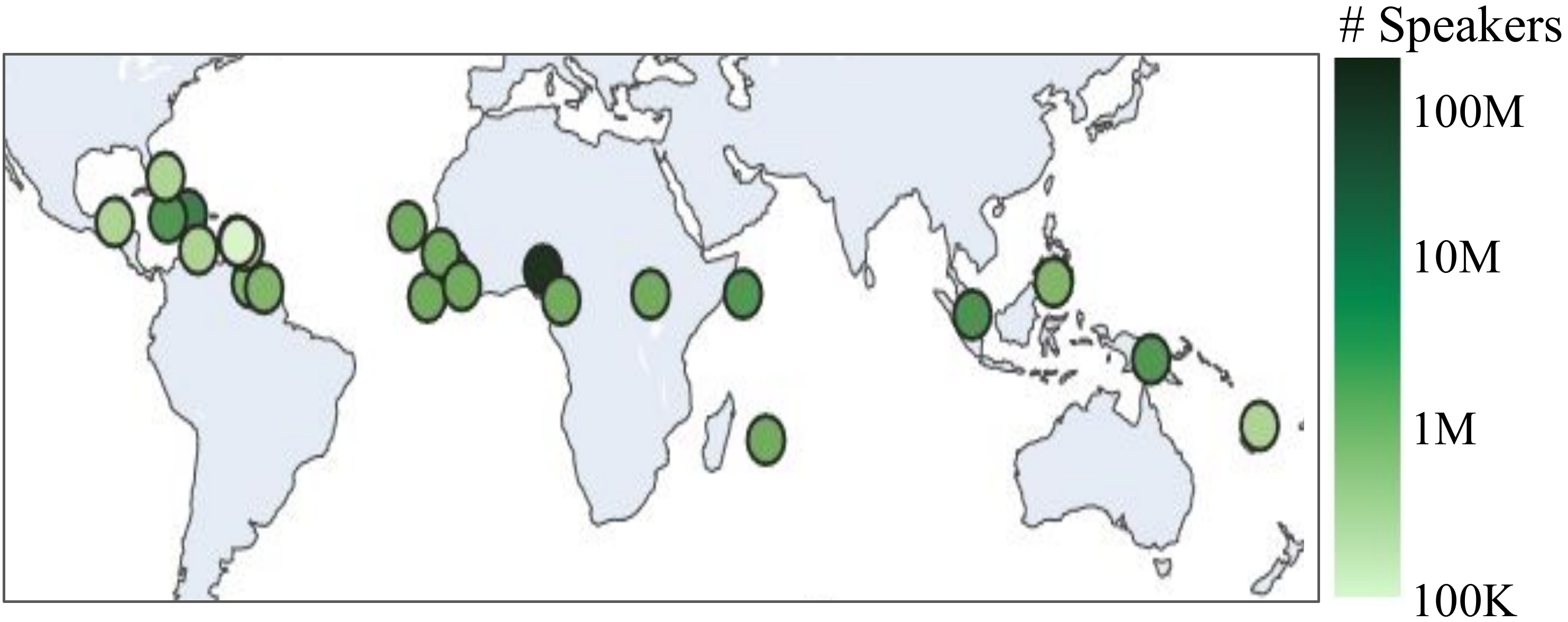}
    \caption{Creoles with a minimum of a hundred thousand speakers are shown here (Hawaiian Pidgin not pictured). Approximately 180 million creole speakers are represented in this map. Data extracted from \url{https://en.wikipedia.org/wiki/List_of_creole_languages}.}
    \label{fig:figures/creolemap}
\end{figure}

A creole language arises if a {\em pidgin},\footnote{A pidgin is a grammatically simplified language that develops between two or more groups that do not have a language in common. Both pidgins and creoles are sometimes referred to as {\em contact} languages.} developed by adults for use as a second language, becomes the native and primary language of their children. Although a large portion of creole languages have their roots in Western European colonialism and slavery, creole languages still serve as important {\em lingua franca} in multi-ethnic and multilingual communities, and creoles are often an important part of the local identity. 
Moreover, there are more than a hundred million speakers of creole languages world wide (\cref{fig:figures/creolemap}), with similar needs for technological assistance, and yet creoles are still largely absent from NLP research~\cite{joshi-etal-2020-state}.
Haitian Creole, for example, has 9.6 million speakers as of today; Nigerian Pidgin English has 100 million speakers, and Singaporean Colloquial English (Singlish) has 3.5 million speakers. 
This paper sets out to collect existing resources for these three languages and provides language models for them. 
In doing so, we wish to take the nature of creole languages into account, not necessarily assuming that our best approaches to modeling non-creole language are also best for the creole languages.

\begin{figure*}[t]
    \centering
    \includegraphics[scale=.45]{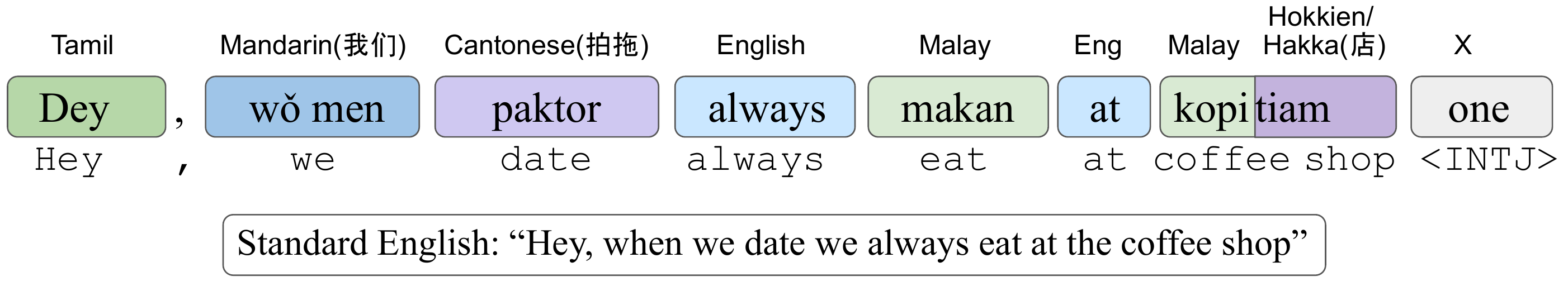}
    \caption{Example sentence in Singlish featuring multilingual vocabulary, Chinese-style topic prominence combined with a subordinate clause with English word order, and a final interjection representing a discourse particle; a common feature of Singlish. Example from \url{https://languagelog.ldc.upenn.edu/nll/?p=25758}.}
    \label{fig:figures/singlishexample}
\end{figure*}

The nature of creole languages has been a matter of much debate in linguistics during the last decade \cite{Sessarego2020NotAG}: Some see creole languages as natural stages in language change cycles \citep{aboh15emergence}, while others see them as a distinct typological class with unique characteristics, including, for example, a very simple morphology \citep{mcwhorter98}. Another feature of creoles is that they exhibit significant variation across groups of speakers \citep{Patrick1999}. \citet{winford99variation} goes as far as to call creoles a {\em continua that cannot be captured under a single grammar}. 

Consider the following pair of sentences from \citet{DBLP:journals/corr/BajpaiPHC17}: 

\begin{enumerate}[noitemsep,topsep=1pt]
    \item[(1)] John sibei hum sup one.
    \item[(2)] John very buaya sia.
\end{enumerate}

Here, according to the authors, both sentences are valid utterances in Singlish, and they both mean {\em John is so lecherous}, but the first would more likely come from a speaker of Chinese, and the second from a Malay speaker.
From this,\footnote{Creole languages clearly differ though in the dynamics that affect their drift. For example, \citet{yakpo_2021} discuss two seemingly similar creole languages, Krio (Sierra Leone) and Pichi (Equatorial Guinea). Both creoles have English as their lexifier, but while Krio is spoken alongside English, Pichi is spoken alongside Spanish. The two creoles, as a consequence, exhibit a clear difference. Krio has converged increasingly toward English, while Pichi has neither converged toward English nor Spanish.} we derive the conjecture that creole language models can benefit from learned mixtures of source languages. Training on mixtures of source languages has been applied to language modeling of code-switched language \cite{pratapa-etal-2018-language}, and it is clear from examples such as the one in   \cref{fig:figures/singlishexample} that creole languages, at the sentence level, share commonalities with code-switched language, with vocabularies drawn from multiple source languages. 
To exploit synergies with learned mixtures of source languages, and to obtain robust performance across related, but unseen distributions, we explore ways of training creole language models with distributionally robust objectives \cite{oren-etal-2019-distributionally}. Our results below, however, show that, somewhat surprisingly, this conjecture is probably not true, at least not in a straight-forward way. 

\paragraph{Contributions} We combine existing datasets and present pretrained language models for the following creole languages: Nigerian Pidgin English, Singaporean Colloquial English (Singlish), and Haitian Creole. 
We perform intrinsic evaluation (word prediction), as well as extrinsic evaluation (part-of-speech tagging and named entity recognition). 
Comparing language models trained with empirical risk minimization to languages models trained with robust objectives, we observe that training with multiple related languages does not improve creole modeling; and also, somewhat surprisingly, that models with empirical risk minimization 
are superior to models robust across domains. 
We hence investigate {\em why} this is: in particular, whether it is due to over-parameterization, insufficient regularization~\citep{sagawa19dro}, or relative distributional stability~\citep{NIPS2006_b1b0432c}. 
We observe no significant difference for language models with fewer parameters or higher degree of regularization.
On the other hand, we find that the underlying reason might be the relative stability of the creoles, which show no significant drift.

\section{Related Work}

\paragraph{NLP research on creoles}
Despite the unique features of creoles that make them an interesting application for multilingual and cross-lingual NLP, as well as the open-ended debate about the linguistic nature of creoles \cite{Sessarego2020NotAG}, little attention has been devoted to creoles in NLP. 
(We present the works related to the specific creoles of focus in this paper in \cref{sec:creolesandcorpora}.)  
One relevant work by \citet{Murawaki2016StatisticalMO} explored the typological status of creoles and also introduced a method for statistical modeling of creole genesis. To start, the authors reported that binary SVM classification of creole and non-creole languages failed to distinguish the two classes, even though their underlying distributions are quite different. After this, they introduce a statistical model of creoles, formulated as a mixture of its influential languages and an inferred "restructurer", which is set of possible linguistic feature distributions that are observed across languages included in their experiments. Overall, this work showcases how statistical modeling methods can be useful for investigating the language evolution of creoles, however there is also no discussion of how their findings could help others extend current NLP methods for creoles.  

\paragraph{NLP research on pidgins and code-switching} Creoles are pidgins that have consolidated over time to become a first language for new generations of speakers. The NLP literature on pidgins is even more sparse than the literature on creoles, because many pidgins that did not undergo creolization have gone extinct, such as Maritime Polynesian Pidgin \cite{2014PidginsandCreoles}.  
Code-switching literature, however, is also relevant, as both pidgins and creoles also draw from other languages. Importantly, pidgins differ from code-switching or mixed language in that code-switching typically only occurs between two bilingual or highly proficient speakers of two languages. Pidgins, on the other hand, are derived from multiple languages, and spoken by those who do not fluently speak every language involved. The NLP literature on code-switching is surprisingly rich, however. We refer readers to \citet{cetinoglu-etal-2016-challenges} and \citet{dogruoz-etal-2021-survey} for an overview. 

\paragraph{Computational research on language evolution} Research on creoles is more common in the field of language evolution than in NLP. In particular, work on creoles in this field typically focuses on their computational modeling, their emergence \cite{nakamura09}, and their evolution \cite{ModelingtheEvolutionofCreoles, Furman2020EvolvingAA}.
Other creole modeling efforts in this space may be more tailored towards specific linguistic insights~\cite{Parkvall2008TheSO}. 
While these studies demonstrate that work on creoles is being done in a computational space, it is difficult to apply conclusions from them to NLP, because distinct empirical assumptions are made in these two research areas.

\paragraph{Distributionally robust optimization}
Effectively learning to model and predict underrepresented subdistributions has always been a challenge in machine learning, e.g., when predicting rare classes,~\citep{scheireretal, fei-liu-2016-breaking} or classes of examples from rare domains \citep{zheng2020outofdomain} or minority groups \cite{pmlr-v80-hashimoto18a}.
Often, underrepresented data is ignored or learned poorly by the models \cite{NEURIPS2020_1e14bfe2}, compared to their over-represented counterparts. 
Distributionally Robust Optimization (DRO) \cite{pmlr-v80-hashimoto18a,sagawa19dro} aims to minimize the loss on {\em all} sub-populations, rather than minimizing their average \cite{BenTal2013RobustSO}.
DRO has been particularly useful in the domain of algorithmic fairness \citep{pmlr-v80-hashimoto18a}, but has also been found to boost performance on underrepresented domains in language modeling  \citep{oren-etal-2019-distributionally} and is generally applicable in situations with drift \cite{koh2021wilds}. 

\section{Creoles and Corpora}
\label{sec:creolesandcorpora}
While creole languages are spoken by hundreds of millions, and are often a {\em lingua franca} within a larger community, only a handful of resources exist for creoles presently. 
Some challenges to collecting data resources for creole languages can be a creole's non-standardized orthography, e.g. Haitian Creole \cite{hewavitharana-etal-2011-cmu}, or the specific contexts in which creoles are used -- it may not always be used in official capacities for news, education, and official documents, even if the creoles are widely used in most other aspects of life \cite{shah-sanghavi17}.
This of course complicates data collection. 
In this work, we focus on the following creoles, as they each have diverse linguistic makeup and have {\em some} existing datasets: 

\paragraph{Nigerian Pidgin English}
West Africa is one of the world's most linguistically diverse places, with Nigeria alone having over 400 languages \cite{UFOMATA1999315}. 
Recent work to advance African NLP has led to the creation of several datasets in Nigerian Pidgin English~\citep{agic-vulic-2019-jw300, Ogueji2019PidginUNMTUN, NdubuisiObi2019WetinDW, caron-etal-2019-surface, oyewusi2021naijaner, adelani2021masakhaner, Oyewusi2020SemanticEO}, 
which makes it particularly well-resourced in comparison to other creole languages. 
Nigerian Pidgin English, also referred to as simply Nigerian Pidgin, can further be understood as a member in the larger family of West African Pidgins, as many West African countries have their own unique variation of this creole, but all share influences from many of the same languages, such as Igbo, Hausa, and Yoruba. 

The first sizeable Nigerian Pidgin dataset comes from \citet{agic-vulic-2019-jw300}, who collected parallel text from several magazines written by a religious society, which have parallel translations in many languages. 
This dataset has been utilized in the first attempts to develop baselines for machine translation of Nigerian Pidgin English \cite{Ogueji2019PidginUNMTUN, Ahia2020TowardsSA}. Furthermore, \citet{Ogueji2019PidginUNMTUN} also introduced the first corpus of Nigerian Pidgin English to further facilitate machine translation from Nigerian Pidgin into English. 
\citet{NdubuisiObi2019WetinDW} also introduced a code-switching corpus of news articles and online comments in both Nigerian Standard English and Nigerian Pidgin. 
In this work, they discuss some challenges of working with Nigerian Pidgin, such as non-standardized spelling. 
They also find that different topics prompt code-switching to Nigerian Pidgin over Nigerian Standard English.
More task-specific Nigerian Pidgin datasets have been introduced for Universal Dependency Parsing \cite{caron-etal-2019-surface}, named entity recognition \cite{oyewusi2021naijaner, adelani2021masakhaner}, sentiment analysis \cite{Oyewusi2020SemanticEO}, and speech recognition \cite{Bigi2017DevelopingRF, Ajisafe2020TowardsET}.

\paragraph{Singlish}
Singaporean Colloquial English, also known as {\em Singlish}, has English as a source language, but also draws parts of its grammar and vocabulary from languages such as Mandarin, Cantonese, Hakka, Hokkien, Malay, and Tamil.
Presently, few publicly available datasets exist in Singlish, as this creole is primarily utilized for informal conversation between people and not for official purposes. 
The largest relevant corpus is The National University of Singapore SMS Corpus from \citet{SinglishSMS}, which consists of over 67,000 text messages written by Singaporeans. 
Qualitatively, we observed that this dataset is much closer to Standard English, albeit with noise from outdated SMS language, than the example provided in \cref{fig:figures/singlishexample}, but, within this data, we still observe many hallmark features of Singlish such as discourse markers and vocabulary from relevant languages. 
 \citet{tan-etal-2020-mind} have also released a webcrawler that collects posts from an popular Singaporean forum about hardware, where discussion is often in Singlish. They use the resulting Singlish corpus as part of their work to investigate the role of inflection for NLP with non-standard forms of English.
Beyond plain text corpora, \citet{wang-etal-2017-universal} introduced the first Singlish Universal Dependency dataset, which was further expanded upon in \citet{WangSinglishUD}. \citet{chau-etal-2020-parsing} used this dataset as a low-resource language test case for their method of pretraining mBERT~\cite{devlin-etal-2019-bert}. 
Finally, a few studies have been done on private datasets for sentiment analysis \cite{DBLP:journals/corr/BajpaiPHC17, SinglishSenticNet}, and polarity detection \cite{SenglishSenticPatterns}.

\begin{table}[t]
\centering
\setlength{\tabcolsep}{1.8pt}
\renewcommand{\arraystretch}{1.0}
\small
\begin{tabular}{lll}
    \toprule
    \multicolumn{1}{l}{\textbf{Language}} & \textbf{Source} & \textbf{Domain} \\ 
    \midrule
    en, fr, es, pt, yo, zh, ta & WMT-News 2020 & news \\ 
    \\[-.8em]
    ms & Malay 30k News & news \\ 
    \\[-.8em]
    Nigerian Pidgin & PidginUNMT Corpus & news \\
    \\[-.8em]
    Singlish & Singapore SMS Corpus & sms \\
    \\[-.8em]
    Haitian Creole & Disaster Response Corpus & sms \\ 
\bottomrule
\end{tabular}
\caption{Data resources utilized in our experiments.}
\label{tab:datapool}
\end{table}
\paragraph{Haitian Creole}
Haitian Creole exhibits a combination of French with many West African languages (e.g. Igbo, Yoruba, Fon, etc.). 
Haitian Creole seized the attention of the machine translation community in the aftermath of the 2010 earthquake crisis in Haiti, during which \citet{Munro10crowdsourcedtranslation,journals-ir-Munro13} developed the Haitian Disaster Response Corpus.
This is a parallel Haitian--English dataset of SMS messages related to the crisis, to enable rapid development of machine translation systems to assist the crisis response. 
This dataset was included in the 2011 Workshop for Machine Translation~\cite{WMT:2011}, in conjunction with data from the medical domain, newswire, and a Haitian glossary.\footnote{\url{http://www.speech.cs.cmu.edu/haitian/text/}.}
Several studies used this data to extend methods in statistical machine translation~\cite{Hu2011TheVO, hu-EtAl:2011:WMT, rcostajussa-banchs:2011:WMT} as well as spell checking and data cleaning \cite{ stymne:2011:WMT}.

\begin{table}[t]
\centering
\setlength{\tabcolsep}{1.8pt}
\renewcommand{\arraystretch}{1.0}
\small
\begin{tabular}{llrrr}
    \toprule
    \textbf{Creole} & \textbf{Langs} & \textbf{\begin{tabular}[c]{@{}c@{}}\# Train\\ Mixed-Lang\end{tabular}} & \textbf{\begin{tabular}[c]{@{}c@{}}\# Train\\ Creole-Only\end{tabular}} & \textbf{\begin{tabular}[c]{@{}c@{}}\# Dev\\ Creole-Only\end{tabular}} \\ 
    \midrule
    \begin{tabular}[c]{@{}l@{}}Nigerian\\ Pidgin\end{tabular} & \begin{tabular}[c]{@{}l@{}}en, pt,\\ yo\end{tabular} & 230,105 & 53,006 & 3,359 \\ \midrule
    Singlish & \begin{tabular}[c]{@{}l@{}}en, zh,\\ ms, ta\end{tabular} & 265,030 & 67,615 & 2,790 \\ 
    \midrule
    \begin{tabular}[c]{@{}l@{}}Haitian\\ Creole\end{tabular} & \begin{tabular}[c]{@{}l@{}}fr, yo,\\ es\end{tabular} & 32,768 & 8,192 & 988 \\ 
    \bottomrule
\end{tabular}
\caption{Creoles, their influential languages (Langs), and the number of examples in the Train-Dev split for our \textsc{Mixed-Language} and \textsc{Creole-Only} experiments. Both use the same creole-only dev dataset.}
\label{tab:split}
\end{table}

\section{Datasets for Creole Language Models}
\label{sec:ourdata}
We experiment with training language models for creoles with a mixture of creole data, and additional data from languages influential to each creole. 
\paragraph{Data splits}
We begin with the creole datasets noted in \cref{tab:datapool}, and combine them with data of other higher-resource languages that have been influential to the creole. We combine a fixed number of these examples into a \textsc{Mixed-Language} dataset, as described in \cref{tab:split}. The {\sc Mixed-Language} dataset for each creole includes information about the original language of each sentence, so that we can form language-specific groups for DRO (see \cref{subsec:training} for more details on DRO grouping). The total number of train and development examples were determined by the number of sentences in the base (creole) dataset for a 95-5 train-development split. Singlish had equal representation of each language, with 53,006 examples per language, including Singlish.
Haitian Creole  also had equally represented languages, with 8,192 examples for Haitian and each additional language.
For the Nigerian Pidgin {\sc Mixed-Language} dataset, English, Portuguese, and Nigerian Pidgin were composed equally with 67,615 examples each, and Yoruba with only 27,260 examples due to the small size of the original data. Thus, we included 95\% of the Yoruba WMT-News 2020 dataset.

\paragraph{Language identification within creoles}
\label{subsec:langcollection}
As we will see in \cref{sec:results}, training the language models on the {\sc Mixed-Language} dataset with DRO fails to produce positive results. Following from this, we also create a \textsc{Creole-Only} dataset, composed of only the creole examples. In order to sort the creole examples into distinct groups for DRO, we label each creole example by the \textit{collection} of the selected languages present in the sentences, as determined by a language identification algorithm.\footnote{\url{https://fasttext.cc/blog/2017/10/02/blog-post.html}.} Consider the following examples from their respective {\sc Creole-Only} datasets:
\begin{itemize}[wide=0.0em]
    \item[\textbf{Singlish:}] \textit{"treat him makah lah"}\\[-0.1em]
    $~~~$en: 88.19\%, ms: 4.34\%, ta: 0.04\%, and zh: 0.01\% 
    \item[\textbf{Nigerian Pidgin:}] \textit{"Pikin wey like to play wit wetin no dey common and sabi one particular subject reach ground"}\\[-0.1em]
    $~~~$en: 87.46\%, pt: 0.23\%, and yo: 0.03\%
    \item[\textbf{Haitian Creole:}] \textit{"Infomation sou kestion te
    tranble a ak lekol"}\\[-0.1em]
    $~~~$fr: 3.50\%, es: 0.08\%, and yo: 0.01\%
\end{itemize}
While the language identification algorithm is not perfect, the confidence scores for the languages still reflect the high-level trends for the creole examples, namely, that English and Malay (\textit{"makan"}) are indeed present in the Singlish sample, and also that English and Portuguese (\textit{"pikin", "sabi"}) are present in the Nigerian Pidgin example. 
However, for the Haitian Creole example, we see that none of our chosen languages have very high scores from the language identification algorithm, which begs the question: were there other languages with higher confidence from the language identification algorithm? 

\begin{figure}
\centering
\begin{subfigure}{.15\textwidth}
  \centering
  \includegraphics[width=\linewidth, trim={0cm 0cm 0cm 0cm}, clip]{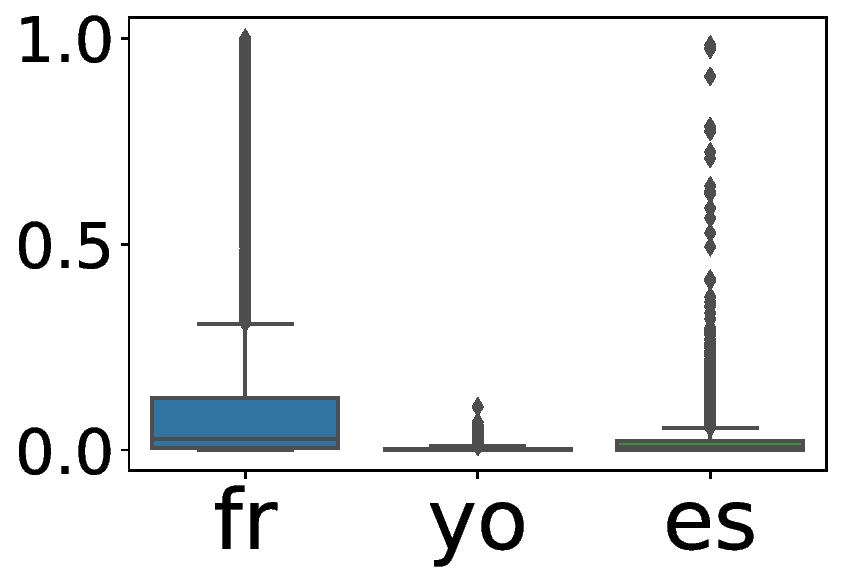}
  \caption*{$~~~~$Haitian (ours)}
  \label{fig:vqa_perm}
\end{subfigure}%
\begin{subfigure}{.15\textwidth}
  \centering
  \includegraphics[width=\linewidth, trim={0cm 0cm 0cm 0cm}, clip]{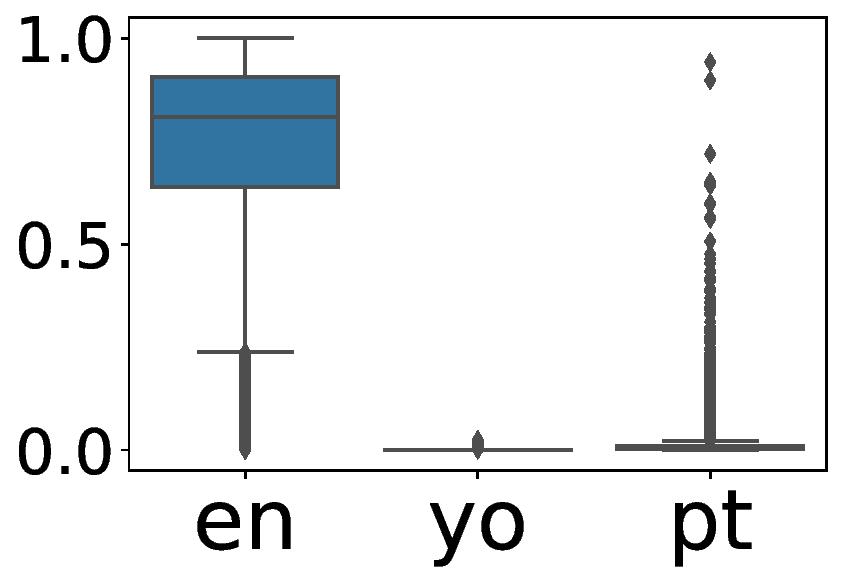}
  \caption*{$~~~~$Nigerian (ours)}
  \label{fig:ref_perm}
\end{subfigure}%
\begin{subfigure}{.15\textwidth}
  \centering
  \includegraphics[width=\linewidth, trim={0cm 0cm 0cm 0cm}, clip]{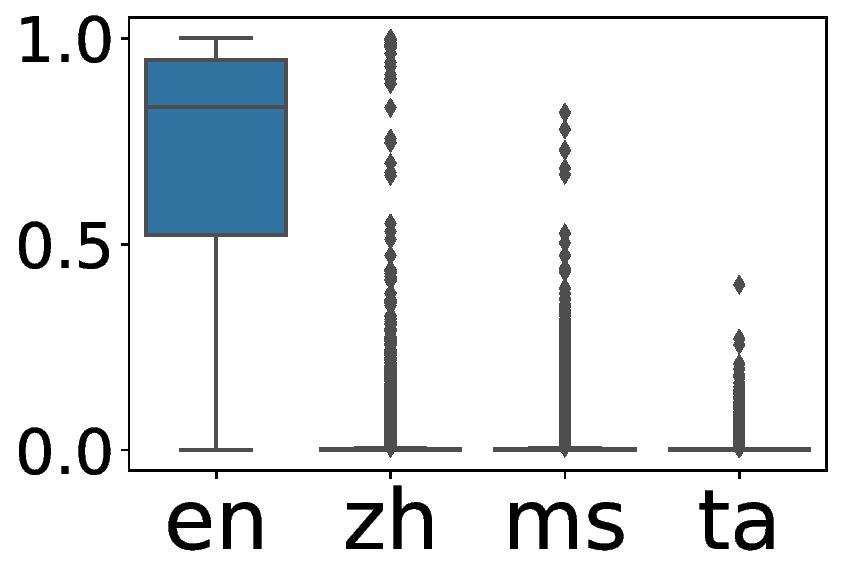}
  \caption*{$~~~~$Singlish (ours)}
  \label{fig:nlvr_perm}
\end{subfigure}
\begin{subfigure}{.15\textwidth}
  \centering
  \includegraphics[width=\linewidth, trim={0cm 0cm 0cm 0cm}, clip]{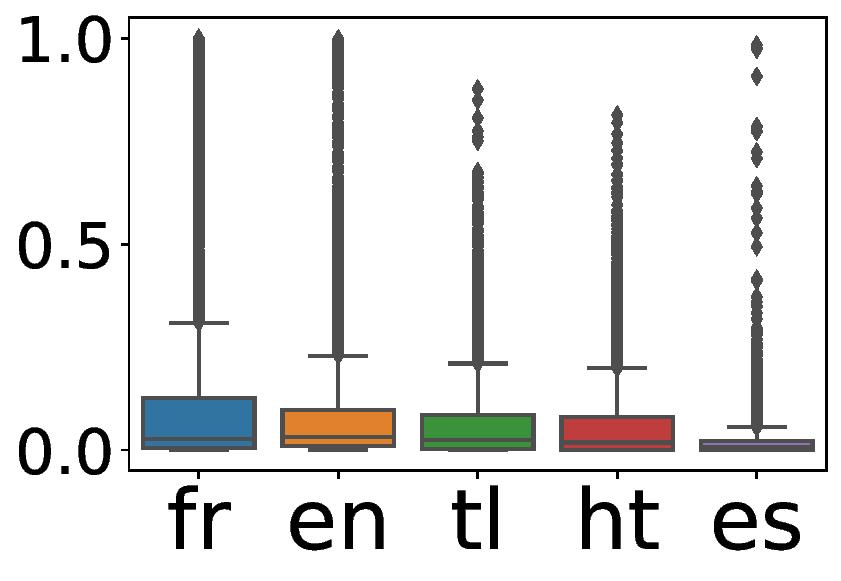}
  \caption*{$~~~~$Haitian (top-5)}
  \label{fig:ir_perm}
\end{subfigure}%
\begin{subfigure}{.15\textwidth}
  \centering
  \includegraphics[width=\linewidth, trim={0cm 0cm 0cm 0cm}, clip]{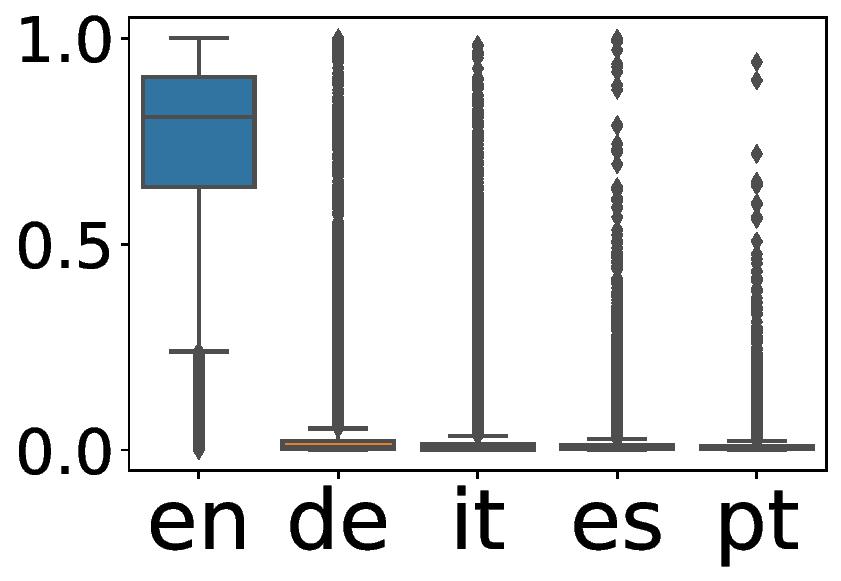}
  \caption*{$~~~~$Nigerian (top-5)}
  \label{fig:sub2}
\end{subfigure}
\begin{subfigure}{.15\textwidth}
  \centering
  \includegraphics[width=\linewidth, trim={0cm 0cm 0cm 0cm}, clip]{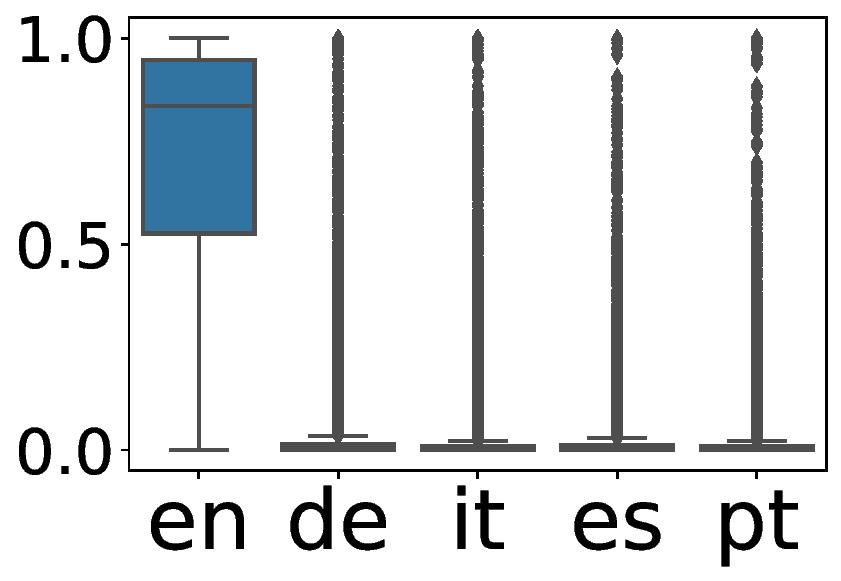}
  \caption*{$~~~~$Singlish (top-5)}
  \label{fig:sub2}
\end{subfigure}
\caption{Distributions of identified languages across the \textsc{Creole-Only} test set. \textbf{Top:} distributions for the influential languages included in \textsc{Mixed-Language}. \textbf{Bottom:} distributions of the five languages that had the highest prediction scores for each creole, where we see a bias towards European languages.}
\label{fig:perm}
\end{figure}

To ensure that the languages we chose are well-represented in the creole examples, we looked at the distribution of the identified languages across examples in our {\sc Creole-Only} datasets in \cref{fig:perm}. From this, we observe that choosing to identify languages specifically related to the creole (i.e. the same languages we included in the {\sc Mixed-Language} datasets) is more reliable than trusting the language identifier pick the top five languages with the highest confidence -- there appears to be a bias for falsely predicting European languages, even on creole data unrelated to these languages, as well as some strange outliers, such as Tagalog being the third most commonly predicted language for Haitian Creole sentences. Also, we see that Haitian Creole itself was a commonly identified language, which could explain the low confidence scores for French and Spanish in the example above. 
Finally, among our specifically chosen languages for the creoles, we see that, although the source language (e.g. English or French) is most dominant, the other languages are still well distributed, with the exception of Yoruba. We surmise that the densely distributed, low-confidence scores for Yoruba can probably be attributed to the fact that Yoruba is a lower-resourced language.

\section{Experiments}
In this section, we detail our experimental setups.
We make our code and models publicly available.\footnote{\url{https://github.com/hclent/creole-dro}}

\subsection{Training} 
\label{subsec:training}
Using the datasets described above, we conduct several experiments to assess how different training strategies affect the modeling of creoles.
We conduct all the experiments on both English BERT and multilingual mBERT models~\cite{devlin-etal-2019-bert}.
As our baseline, we consider pretrained BERT$_\text{Base}$ and mBERT models, and evaluate them on our development splits for the creoles. We then assess the effectiveness of two popular training strategies: Empirical Risk Minimization (ERM) and Distributionally Robust Optimization (DRO). In this case, ERM consists of masked language modeling over all the data points in each dataset, in a similar fashion as done during pretraining.

For DRO, we utilize the WILDS library \citep{DBLP:journals/corr/abs-2012-07421}, which uses metadata associated with the input data to form the groups for DRO. 
In our case, we investigate three grouping strategies: grouping with language information as metadata ({\bf DRO-Language}), as well as with two additional control experiments. 
In the first control experiment, we assign all training examples to the same group ({\bf DRO-One}), such that that DRO is optimizing over only one large group.
In the second control experiment, we randomly assign examples to one of four groups ({\bf DRO-Random}). 
The motivation of for these control experiments is to ensure that improvements for DRO are actually grounded in the language information, and not an artifact of the WILDS grouping algorithm.

In {\bf DRO-Language}, information about the examples' language makeup is used to determine the groups. In {\sc Mixed-Language}, we rely on our knowledge of where the examples were sampled from, but in {\sc Creole-Only}, we subdivide the creole examples depending on their etymology. Specifically, grouping is done as follows in our two data setups outlined in \cref{sec:ourdata}:
\begin{itemize}[noitemsep,topsep=1pt]
    \item \textbf{\textsc{Mixed-Language}:} Here, grouping is done over the languages in the training data. For example, in the case of Nigerian Pidgin, if a sentence originally comes from the Yoruba corpus, it is assigned to the Yoruba group, and similarly for Nigerian Pidgin and the other languages listed in \cref{tab:split} for each creole.

    \item \textbf{\textsc{Creole-Only}:} Here, as we only have the creole samples, grouping is done over the confidence scores from the collection of the influential languages (see Section \cref{subsec:langcollection}). An example is assigned to one of $2^{N}$ groups, representing the combinations of detected languages in a sentence. $N$ is the number of languages listed in \cref{tab:split} (Langs) for each creole, and presence of a language is derived from its confidence score by the language identifier: if there is a confidence of 0.1\% or higher that the language is represented in the sentence, then it is considered as present. 
\end{itemize}

\begin{table*}[t]
\centering
\small
\begin{tabular}{clrrr rrrrr rrrr}
\toprule
& && \multicolumn{3}{c}{\textbf{Nigerian Pidgin}} && \multicolumn{3}{c}{\textbf{Singlish}} && \multicolumn{3}{c}{\textbf{Haitian Creole}} \\
\cmidrule(lr){4-6} \cmidrule(lr){8-10} \cmidrule(lr){12-14}
&\textbf{BERT} && \multicolumn{1}{c}{P@1} & \multicolumn{1}{c}{P$_\text{D}$@1} & \multicolumn{1}{c}{PLL} && \multicolumn{1}{c}{P@1} & \multicolumn{1}{c}{P$_\text{D}$@1} & \multicolumn{1}{c}{PLL} && \multicolumn{1}{c}{P@1} & \multicolumn{1}{c}{P$_\text{D}$@1} & \multicolumn{1}{c}{PLL} \\
\midrule
&Pretrained && 22.79 & 10.92 & {142.65} && 23.94 & {21.09} & {76.01} && 18.84 & {5.65} & 177.40 \\
\midrule 
\multirow{4}{*}{\rotatebox[origin=c]{90}{\sc Mixed}}&ERM && \textbf{63.83} & {\textbf{59.97}} & \textbf{42.41} && \textbf{46.77} & {\textbf{42.89}} & \textbf{41.06} && \textbf{68.09} & {\textbf{43.35}} & \textbf{55.04} \\
\cmidrule{2-14}
&DRO-One && 60.99 & {56.76} & {52.51} && 44.23 & {40.73} & {49.18} && 57.04 & {36.73} & 121.51 \\
&DRO-Random && 60.40 & {56.33} & {52.69} && 43.33 & {39.07} & {49.14} && 57.65 & {36.16} & 119.17 \\
&DRO-Language && 60.40 & {54.80} & {54.17} && 43.19 & {39.57} & {48.88} && 57.55 & {36.69} & 118.85 \\
\midrule 
\multirow{4}{*}{\rotatebox[origin=c]{90}{\sc C-Only}}&ERM && \textbf{73.72} & {\textbf{71.38}} & \textbf{28.14} && \textbf{53.80} & {\textbf{51.26}} & \textbf{34.22} && \textbf{73.15} & {\textbf{55.50}} & \textbf{55.51} \\
\cmidrule{2-14}
&DRO-One && 64.28 & {59.86} & {61.81} && 45.34 & {43.59} & {66.53} && 58.16 & {36.91} & 144.46 \\
&DRO-Random && 63.72 & {59.31} & {60.31} && 45.73 & {42.40} & {64.16} && 57.65 & {37.41} & 142.04 \\
&DRO-Language && 63.58 & {59.74} & {56.82} && 44.73 & {40.57} & {53.72} && 56.94 & {35.50} & 138.60 \\
\bottomrule
\end{tabular}
\caption{Intrinsic evaluation:  Precision@1 (P@1), Precision@1 for words in our creole dictionary ({P$_\text{D}$@1}), and average Pseudo-log-likelihood score (PLL). We report results for {\sc Mixed-Language} (top) and {\sc Creole-Only} (bottom). We note that ERM consistently outperforms the language models trained with robust objectives.}
\label{tab:mixedlang_results}
\end{table*}

\subsection{Evaluation}
We perform two types of evaluation: intrinsic -- based on the MLM training objective -- and extrinsic -- on traditional downstream NLP tasks.

\paragraph{Intrinsic evaluation}
We evaluate our language models intrinsically with the following metrics:
\begin{itemize}[noitemsep,topsep=1pt]
    \item \textbf{Precision at $k$ (P@$k$):} Precision of the language model in predicting a random masked token per sentence. This allows us to assess the general performance following the training objective. In the following, we report P@1. Results at $k=\{5,10\}$ are in the App.
    \item \textbf{Dictionary-based precision at $k$ (P$_\text{D}$@$k$):} Due to their nature, most of the words in a creole sentence are from the corresponding source language (see \cref{fig:perm}). Hence, for a more principled measurement of precision, we collect online dictionaries of our creoles.\footnote{Nigerian Pidgin: \url{http://naijalingo.com/}.\\ Singlish: {\url{http://www.mysmu.edu/faculty/jacklee/}}.\\ Haitian Creole: \url{https://kreyol.com/dictionary.html}.} We perform the same MLM task as above, but this time only mask words belonging to the creole dictionaries. By doing so, we can obtain a more accurate measure of what the LMs have learned. We again report results at $k=1$ here, and refer the reader to the App. for $k=\{5,10\}$.
    \item \textbf{Mean pseudo-log-likelihood score (PLL):} Following recent studies~\cite{pmlr-v101-shin19a,wang2019bert,salazar-etal-2020-masked}, we measure the pseudo-log-likelihood scores from MLMs given by summing the conditional log probabilities $\log \mathbb{P}_\text{MLM}(w_t|\mathbf{w}_{\text{\textbackslash} t})$ of each token $w_t$ in a sentence $\mathbf{w}=\langle w_1, \dots, w_T \rangle$. These are obtained in BERT by replacing $w_t$ with the special \texttt{[MASK]} token. Here, we report the mean score given by:
    \begin{equation}
    \resizebox{.44 \textwidth}{!}{
    $
        \text{PLL} = \dfrac{1}{|\mathcal{C}|}\sum\limits_{{\mathbf{w}\in\mathcal{C}}}\dfrac{1}{|\mathbf{w}|}\sum\limits_{w_t\in\mathbf{w}}\log \mathbb{P}_\text{MLM}(w_t|\mathbf{w}_{\text{\textbackslash} t}; \theta),
    $
    }
    \end{equation}
    where $\mathcal{C}$ denotes the evaluation corpus, and $\theta$ denotes a model’s parameters.
\end{itemize}

\paragraph{Extrinsic evaluation}
We also perform an extrinsic evaluation of our models on downstream tasks, for the datasets that are available. 
Specifically, we train and evaluate models for Nigerian Pidgin NER and POS tagging with Universal Dependencies \citep[UPOS]{nivre20universal}, as well as Singlish UPOS. We fine-tune our pretrained language models on the training sets of these two tasks and evaluate them on the corresponding test sets. 

\subsection{Framework}
We write our code in PyTorch~\cite{pytorchpaper}. 
In particular, for language model training, we rely on the HuggingFace Transformers library~\cite{transformers_paper}, and the WILDS library~\cite{DBLP:journals/corr/abs-2012-07421} for DRO.
Models are fine-tuned for 100,000 steps with batch size of 16.
For downstream tasks, we use MaChAmp \citep{van-der-goot-etal-2021-massive} and train our models for $10$ epochs.
The best checkpoints were selected based on performance on the dev sets.
Unless otherwise specified, we use the default hyperparameters.
Our experiments are run on one NVIDIA TitanX GPU in a shared cluster.

\section{Results and Analyses}
\label{sec:results}

\paragraph{Intrinsic evaluation}
The main finding of the intrinsic evaluation is that ERM outperforms DRO for all grouping strategies across all metrics. 
We also observe that P$_\text{D}$@k is a more difficult task than the standard precision at $k$, with randomly masked tokens (see \cref{sec:appendix} for full results with both BERT and mBERT). 
Moreover we find that the DRO models often have a much higher perplexity than ERM.  
Finally, the results show that, between the {\sc Mixed-Language} and {\sc Creole-Only} experiments, the latter performed better, demonstrating that training on additional data was not useful for learning language models for creoles. 
While we only report results for BERT here, we observe the same patters with mBERT (see \cref{sec:appendix}).

\paragraph{Extrinsic evaluation}
Here, we observe the same trend as in the intrinsic evaluation: ERM performs better than DRO (see \cref{tab:downstream}). Although for Nigerian Pidgin DRO-Language performs better than ERM on both NER and UPOS, the gap between the scores is too small to draw concrete conclusioins from.

\begin{table}[t]
\centering
\setlength{\tabcolsep}{1.8pt}
\renewcommand{\arraystretch}{1.0}
\small
\begin{tabular}{clccc}
    \toprule
    && \multicolumn{2}{c}{\textbf{Nigerian Pidgin}} & \multicolumn{1}{c}{\textbf{Singlish}} \\ 
    \cmidrule(lr){3-4} \cmidrule(lr){5-5}
    &\textbf{BERT} & NER [F$_1$] & UPOS [Acc] & UPOS [Acc] \\
    \midrule
    \multirow{2}{*}{\rotatebox[origin=c]{90}{\sc\tiny Mixed}}&ERM & 87.86 & {98.00} & {\bf 91.24} \\
    &DRO-Language& {\bf 88.40} & {\bf 98.06} & 90.22 \\ 
    \midrule
    \multirow{2}{*}{\rotatebox[origin=c]{90}{\sc\tiny C-Only}}&ERM & {\bf 87.98} & {\bf 98.04} & {\bf 91.17} \\
    &DRO-Language & 87.12 &{97.98} & 90.44 \\
    \bottomrule
\end{tabular}
\caption{Extrinsic evaluation. Similar performance on downstream tasks across all models demonstrate show that language model training did {\em not} benefit significantly from neither DRO nor data in related languages.}
\label{tab:downstream}
\end{table}

\paragraph{}
There are several factors that could have influenced the DRO models to perform worse than ERM. We explore their effects below. 

\paragraph{Over-parameterization}
Over-parameterization is known to be problematic for DRO \cite{sagawa19dro}. In order to investigate the role of over-parameterization in our experiments, we ran additional {\sc Mixed-Language} experiments on Nigerian Pidgin English, with different sized BERT models, namely BERT$_\text{Tiny}$, BERT$_\text{Small}$ {\cite{jiao-etal-2020-tinybert}}, and BERT$_\text{Base}$. The results in~\cref{tab:overparam} demonstrate that over-parameterization was not a leading cause for DRO failure, otherwise we would expect for smaller BERT versions to have relative better performance compared to the corresponding ERM runs. Instead, we see that standard BERT works fine for this task, and over-parameterization is not the cause of poor performance of DRO in our experiments.

\begin{table}[t]
\centering
\renewcommand{\arraystretch}{1.0}
\small
\begin{tabular}{llrrr}
    \toprule
    && \multicolumn{3}{c}{\textbf{Nigerian Pidgin}} \\ 
    \cmidrule(lr){3-5}
    \textbf{BERT} & \textbf{Size} & P@1 & P$_\text{D}$@1 & PLL \\
    \midrule
    \multirow{3}{*}{ERM} & Tiny & 31.31 & 26.12 & 110.23 \\
     & Small & 47.39 & 46.75 & 77.47 \\
     & Base & 63.83 & 59.97 & 42.41 \\ 
    \midrule
    \multirow{3}{*}{DRO-Language} & Tiny & 31.00 & 23.09 & 99.70 \\
     & Small & 43.00 & 37.75 & 82.50 \\
     & Base & 60.40 & 54.80 & 54.17 \\ 
    \bottomrule
\end{tabular}
\caption{Over-parameterization experiments with {\sc Mixed-Language} Nigerian Pidgin English data. Smaller sized models do not benefit DRO over ERM.}
\label{tab:overparam}
\end{table}

\paragraph{Regularization}
\citet{sagawa19dro} also discuss how lack of regularization lead to problems for DRO, and how increased regularization is necessary for worst-group generalization. To investigate this potential weakness in our experiments, we run additional experiments using BERT$_\text{Small}$ on {\sc Mixed-Language} data for Nigerian Pidgin English, trying different weight decay values in each \cref{tab:regularization}. If our DRO models were suffering from insufficient regularization, we would expect that increasing the regularization factor of weight decay would boost performance. However, we find no meaningful effect of this hyperparameter, which leads us to believe that insufficient regularization is not a driving factor in the underperformance of DRO compared to ERM.

\begin{table}[t]
\centering
\setlength{\tabcolsep}{2.8pt}
\renewcommand{\arraystretch}{1.0}
\small
\begin{tabular}{lcrrr}
    \toprule
    && \multicolumn{3}{c}{\textbf{Nigerian Pidgin}} \\ 
    \cmidrule(lr){3-5}
    \textbf{BERT} & \textbf{Weight Decay} & P@1 & P$_\text{D}$@1 & PLL \\
    \midrule
    ERM & 0.01 & 47.39 & 46.75 & 77.47 \\
    \midrule
    \multirow{4}{*}{DRO-Language} & 0.01 & 43.00 & 37.75 & 82.50 \\
     & 0.05 & 42.86 & 38.47 & 83.03 \\
     & 0.10 & 43.00 & 38.74 & 81.80 \\
     & 0.30 & 42.70 & 39.53 & 81.94 \\
    \bottomrule
\end{tabular}
\caption{Regularization experiments on {\sc Mixed-Language} Nigerian Pidgin data, based on BERT$_\text{Small}$.}
\label{tab:regularization}
\end{table}

\paragraph{Drift and creole stability} Creole languages arise from pidgins, which are initially developed for use as second language. Recent years have seen renewed interest in the classic question of the relationship between pidgin and creole formation and second language acquisition \cite{Plag2009CreolesAI}. 
To investigate the matter of creole stability, we follow \citep{NIPS2006_b1b0432c} and calculate the proxy $\mathcal{A}$-distance (PAD) between different domains of creole data (see ~\cref{tab:a_distance}). 
Specifically, we train an SVM on the BERT encodings.\footnote{Our code is adapted from \url{https://github.com/rpryzant/proxy-a-distance}.} 
Our $\mathcal{A}$-distance results suggest that creole languages do {\em not} exhibit more drift than English when the data are comparable.
This potentially explains why distributionally robust language models do not outperform regular language models trained with empirical risk minimization objectives. 

\begin{table}[t]
\centering
\setlength{\tabcolsep}{1.8pt}
\renewcommand{\arraystretch}{1.0}
\small
\begin{tabular}{llll}
    \toprule
    \textbf{Language} & \textbf{Domain-1} & \textbf{Domain-2} & \textbf{PAD} \\
    \midrule
    English & Disaster Response Corpus & Newswire & 1.75 \\
    Haitian Creole & Disaster Response Corpus & Newswire & 1.47 \\
    \midrule English&EWT-UD&	NUD	&1.04\\
Nigerian&UNMT &NUD&	1.28\\ 
    \bottomrule
\end{tabular}
\caption{Proxy $\mathcal{A}$-distance (PAD) scores on parallel (Haitian) or near-parallel (Nigerian) data. PAD is proportional to domain classification error; hence, large distances mean high domain divergence. Our results suggest that creole languages do {\em not} exhibit significantly more drift than other languages.}
\label{tab:a_distance}
\end{table}

\section{Conclusion}
In this paper, we bring creole languages to the attention of the NLP community. We collect data and train baseline language models for three creoles, and evaluate these models across the downstream tasks of part-of-speech tagging and named entity recognition. Based on previous work suggesting the instability of creole languages \cite{winford99variation,Patrick1999}, we explore the impact of using more robust learning objectives for masked language modeling of creoles, but our results show that vanilla empirical risk minimiziation is superior. We show that this is not the result of over-parameterization or lack of regularization, but instead suggest this is a result of the relative stability of creole languages. We note that it still remains possible that significant improvements could be achieved by modeling dynamics specific to creole languages, i.e., the processes that govern their development, including social factors \cite{holm_2000} and second language acquisition dynamics \cite{Plag2009CreolesAI}.   
 \section{Acknowledgments}
\euflag ~We would like to thank the anonymous reviewers and Samson Tan for their helpful feedback.
We would also like to thank Robert Monarch and Chris Callison-Burch for their help with the Haitian data.
This project has received funding from the European Union’s Horizon 2020 research and innovation programme under the Marie Skłodowska-Curie grant agreement No 801199 (for Heather Lent and Emanuele Bugliarello), the Swedish Research Council Grant 2020-00437 (for Miryam de Lhoneux), and the Google Research Award (for Heather Lent and Anders Søgaard).

\bibliography{custom}
\bibliographystyle{acl_natbib}

\clearpage
\appendix
\section{Appendix} \label{sec:appendix}

\input{appendix/table1}

\clearpage
\input{appendix/table2}

\input{appendix/table3}

\end{document}

%% file: appendix/table1.tex
\begin{table}[h]
\centering
\begin{tabular}{l|l|rrrrrrr}
\toprule
\multicolumn{9}{c}{\textbf{Singlish}}                                                                                                                                                                                                                 \\
\midrule
        Model                                  & Strategy & \multicolumn{1}{l}{P@1} & \multicolumn{1}{l}{P@5} & \multicolumn{1}{l}{P@10} & \multicolumn{1}{l}{PLL} & \multicolumn{1}{l}{P$_{D}$@1} & \multicolumn{1}{l}{P$_{D}$@5} & \multicolumn{1}{l}{P$_{D}$@10} \\
                       \midrule
\multirow{4}{*}{BERT}  & ERM                           & 46.77                   & 68.89                   & 74.34                    & 41.07                   & 42.89                     & 66.76                     & 74.17                     \\
                       & DRO-One       & 44.23                   & 64.73                   & 71.90                    & 49.18                   & 40.73                     & 63.05                     & 70.13                     \\
                       & DRO-Random    & 43.33                   & 65.63                   & 71.58                    & 49.14                   & 39.07                     & 61.02                     & 68.42                     \\
                       & DRO-Language  & 43.19                   & 64.80                   & 71.22                    & 48.88                   & 39.57                     & 61.54                     & 70.34                     \\
                       \midrule
\multirow{4}{*}{mBERT} & ERM                     & 47.78                   & 68.82                   & 75.77                    & 42.26                   & 37.00                     & 61.71                     & 70.85                     \\
                       & DRO-One       & 44.37                   & 65.13                   & 72.54                    & 50.99                   & 33.71                     & 57.79                     & 65.72                     \\
                       & DRO-Random    & 44.34                   & 65.95                   & 72.44                    & 50.39                   & 35.25                     & 59.20                     & 66.93                     \\
                       & DRO-Language  & 43.69                   & 64.91                   & 71.61                    & 50.49                   & 33.45                     & 58.91                     & 67.42                     \\
\bottomrule
\multicolumn{9}{c}{\textbf{Naija}}                                                                                                                                                                                                                    \\
\midrule
                 Model                   &  Strategy                             & \multicolumn{1}{l}{P@1} & \multicolumn{1}{l}{P@5} & \multicolumn{1}{l}{P@10} & \multicolumn{1}{l}{PLL} & \multicolumn{1}{l}{P$_{D}$@1} & \multicolumn{1}{l}{P$_{D}$@5} & \multicolumn{1}{l}{P$_{D}$@10} \\
                       \midrule
\multirow{4}{*}{BERT}  & ERM                            & 63.83                   & 80.52                   & 85.44                    & 42.41                   & 59.97                     & 78.72                     & 83.93                     \\
                       & DRO-One       & 60.99                   & 77.52                   & 82.94                    & 52.51                   & 56.76                     & 76.21                     & 81.94                     \\
                       & DRO-Random    & 60.40                   & 78.44                   & 82.88                    & 52.69                   & 56.33                     & 75.27                     & 81.17                     \\
                       & DRO-Language  & 60.40                   & 77.40                   & 82.69                    & 54.18                   & 54.80                     & 74.48                     & 80.32                     \\
                       \midrule
\multirow{4}{*}{mBERT} & ERM                            & 62.68                   & 80.52                   & 85.55                    & 44.98                   & 62.19                     & 82.25                     & 87.08                     \\
                       & DRO-One       & 60.76                   & 77.74                   & 82.88                    & 58.15                   & 56.43                     & 77.61                     & 82.67                     \\
                       & DRO-Random    & 60.34                   & 77.23                   & 82.24                    & 57.90                   & 56.70                     & 77.50                     & 82.51                     \\
                       & DRO-Language  & 58.88                   & 76.56                   & 81.73                    & 59.91                   & 54.99                     & 76.63                     & 82.50                     \\
\bottomrule
\multicolumn{9}{c}{\textbf{Haitian}}                                                                                                                                                                                                                  \\
\midrule
           Model                   &  Strategy                              & \multicolumn{1}{l}{P@1} & \multicolumn{1}{l}{P@5} & \multicolumn{1}{l}{P@10} & \multicolumn{1}{l}{PLL} & \multicolumn{1}{l}{P$_{D}$@1} & \multicolumn{1}{l}{P$_{D}$@5} & \multicolumn{1}{l}{P$_{D}$@10} \\
                       \midrule
\multirow{4}{*}{BERT}  & ERM                  & 68.09                   & 82.98                   & 87.34                    & 55.05                   & 43.35                     & 63.89                     & 71.35                     \\
                       & DRO-One       & 57.04                   & 71.12                   & 75.58                    & 121.51                  & 36.73                     & 52.55                     & 58.25                     \\
                       & DRO-Random    & 57.65                   & 71.53                   & 75.79                    & 119.17                  & 36.16                     & 50.63                     & 56.38                     \\
                       & DRO-Language  & 57.55                   & 71.23                   & 75.28                    & 118.85                  & 36.69                     & 50.48                     & 55.89                     \\
                       \midrule
\multirow{4}{*}{mBERT} & ERM                           & 60.79                   & 76.70                   & 81.56                    & 60.27                   & 46.35                     & 64.56                     & 70.96                     \\
                       & DRO-One       & 51.06                   & 65.45                   & 69.71                    & 148.12                  & 34.57                     & 49.30                     & 55.61                     \\
                       & DRO-Random    & 50.86                   & 65.05                   & 69.50                    & 146.18                  & 34.52                     & 49.58                     & 55.00                     \\
                       & DRO-Language  & 50.15                   & 64.54                   & 69.40                    & 145.97                  & 33.55                     & 48.21                     & 55.08      \\
\bottomrule
\end{tabular}
\caption{Full results for Mixed-Language experiments.}
\end{table}

%% file: appendix/table2.tex
\begin{table}[t]
\begin{tabular}{l|l|rrrrrrr}
\toprule
\multicolumn{9}{c}{\textbf{Singlish}}                                                                                                                                                                                                                 \\
\midrule
    Model                   &  Strategy  & \multicolumn{1}{l}{P@1} & \multicolumn{1}{l}{P@5} & \multicolumn{1}{l}{P@10} & \multicolumn{1}{l}{PLL} & \multicolumn{1}{l}{P$_{D}$@1} & \multicolumn{1}{l}{P$_{D}$@5} & \multicolumn{1}{l}{P$_{D}$@10} \\
                       \midrule
\multirow{4}{*}{BERT}  & ERM                & 53.80                   & 75.02                   & 80.36                    & 34.22                   & 51.26                     & 74.09                     & 80.15                     \\
                       & DRO-One       & 45.34                   & 64.41                   & 70.14                    & 66.53                   & 43.59                     & 63.42                     & 69.33                     \\
                       & DRO-Random    & 45.73                   & 64.66                   & 71.00                    & 64.16                   & 42.40                     & 64.38                     & 70.74                     \\
                       & DRO-Language  & 44.73                   & 65.16                   & 71.08                    & 57.54                   & 40.57                     & 62.68                     & 69.78                     \\
                       \midrule
\multirow{4}{*}{mBERT} & ERM                         & 56.81                   & 77.03                   & 81.65                    & 34.49                   & 46.87                     & 72.55                     & 79.49                     \\
                       & DRO-One       & 47.49                   & 65.84                   & 70.97                    & 76.57                   & 36.17                     & 56.74                     & 64.51                     \\
                       & DRO-Random    & 47.85                   & 65.88                   & 70.93                    & 74.87                   & 37.66                     & 58.37                     & 65.41                     \\
                       & DRO-Language  & 45.77                   & 64.77                   & 70.39                    & 68.55                   & 33.94                     & 55.01                     & 62.09  \\
\bottomrule
\multicolumn{9}{c}{\textbf{Naija}}                                                                                                                                                                                                                    \\
\midrule
           Model                   &  Strategy                            & \multicolumn{1}{l}{P@1} & \multicolumn{1}{l}{P@5} & \multicolumn{1}{l}{P@10} & \multicolumn{1}{l}{PLL} & \multicolumn{1}{l}{P$_{D}$@1} & \multicolumn{1}{l}{P$_{D}$@5} & \multicolumn{1}{l}{P$_{D}$@10} \\
                       \midrule
\multirow{4}{*}{BERT}  & ERM       & 73.72                   & 88.62                   & 91.99                    & 28.14                   & 71.38                     & 87.33                     & 90.94                     \\
                       & DRO-One       & 64.28                   & 79.37                   & 83.95                    & 61.81                   & 59.86                     & 77.00                     & 81.60                     \\
                       & DRO-Random    & 63.72                   & 79.57                   & 83.92                    & 60.31                   & 59.31                     & 75.55                     & 80.29                     \\
                       & DRO-Language  & 63.58                   & 79.48                   & 84.29                    & 56.83                   & 59.74                     & 77.08                     & 81.84                     \\
\midrule
\multirow{4}{*}{mBERT} & ERM                       & 72.96                   & 87.58                   & 91.15                    & 31.77                   & 70.42                     & 87.58                     & 91.42                     \\
                       & DRO-One       & 63.72                   & 78.36                   & 82.77                    & 76.24                   & 60.78                     & 77.07                     & 81.78                     \\
                       & DRO-Random    & 63.52                   & 77.77                   & 82.18                    & 74.53                   & 61.02                     & 78.01                     & 82.88                     \\
                       & DRO-Language  & 63.13                   & 78.16                   & 82.80                    & 71.77                   & 60.73                     & 77.37                     & 82.25  \\
\bottomrule
\multicolumn{9}{c}{\textbf{Haitian}}                                                                                                                                                                                                                  \\
\midrule
                    Model                   &  Strategy                              & \multicolumn{1}{l}{P@1} & \multicolumn{1}{l}{P@5} & \multicolumn{1}{l}{P@10} & \multicolumn{1}{l}{PLL} & \multicolumn{1}{l}{P$_{D}$@1} & \multicolumn{1}{l}{P$_{D}$@5} & \multicolumn{1}{l}{P$_{D}$@10} \\
                       \midrule
\multirow{4}{*}{BERT}  & ERM                            & 73.15                   & 86.12                   & 88.55                    & 55.51                   & 55.50                     & 71.76                     & 77.94                     \\
                       & DRO-One       & 58.16                   & 71.23                   & 75.48                    & 144.47                  & 36.91                     & 51.39                     & 56.04                     \\
                       & DRO-Random    & 57.65                   & 70.52                   & 75.38                    & 142.04                  & 37.41                     & 52.55                     & 57.72                     \\
                       & DRO-Language  & 56.94                   & 71.33                   & 74.97                    & 138.60                  & 35.50                     & 49.66                     & 55.03                     \\
                       \midrule
\multirow{4}{*}{mBERT} & ERM                       & 66.06                   & 80.45                   & 84.30                    & 69.25                   & 55.58                     & 72.49                     & 78.72                     \\
                       & DRO-One       & 50.35                   & 65.05                   & 69.10                    & 174.45                  & 35.86                     & 51.60                     & 56.72                     \\
                       & DRO-Random    & 48.63                   & 64.03                   & 67.78                    & 172.26                  & 32.54                     & 48.31                     & 53.60                     \\
                       & DRO-Language  & 49.14                   & 64.24                   & 68.69                    & 167.90                  & 34.59                     & 49.34                     & 55.45 \\
                       \bottomrule
\end{tabular}
\caption{Full results for Creole-Only experiments.}
\end{table}

%% file: appendix/table3.tex
\begin{table}[b]
\begin{tabular}{l|l|rrrrrrr}
\toprule
    Dataset                      &   Model    & \multicolumn{1}{l}{P@1} & \multicolumn{1}{l}{P@5} & \multicolumn{1}{l}{P@10} & \multicolumn{1}{l}{PLL} & \multicolumn{1}{l}{P$_{D}$@1} & \multicolumn{1}{l}{P$_{D}$@5} & \multicolumn{1}{l}{P$_{D}$@10} \\
                          \midrule
\multirow{2}{*}{Singlish} & BERT  & 23.94                   & 38.49                   & 45.09                    & 76.01                   & 21.09                     & 36.65                     & 42.22                      \\
                          & mBERT & 14.30                   & 23.12                   & 27.03                    & 92.97                   & 10.23                     & 23.57                     & 29.85                      \\
                          \midrule
\multirow{2}{*}{Nigerian Pidgin}    & BERT  & 22.79                   & 34.04                   & 39.88                    & 142.66                  & 10.92                     & 18.07                     & 22.96                      \\
                          & mBERT & 14.90                   & 26.34                   & 31.87                    & 153.54                  & 8.08                      & 16.24                     & 20.72                      \\
                          \midrule
\multirow{2}{*}{Haitian Creole}  & BERT  & 18.84                   & 30.60                   & 37.59                    & 177.40                  & 5.65                      & 11.89                     & 16.29                      \\
                          & mBERT & 11.96                   & 22.39                   & 27.96                    & 175.14                  & 7.10                      & 12.20                     & 16.76                     \\
\bottomrule
\end{tabular}
\caption{Full results for pretrained baselines.}
\end{table}